\documentclass[conference]{IEEEtran}
\IEEEoverridecommandlockouts
\usepackage{cite}
\usepackage{amsmath,amssymb,amsfonts}
\usepackage{algorithmic}
\usepackage{graphicx}
\usepackage{textcomp}
\usepackage{xcolor}
\usepackage{booktabs}
\usepackage{multirow}
\usepackage{xspace}
\usepackage{balance}
\usepackage{bm}
\usepackage{url}
\usepackage{hyperref}
\usepackage{marvosym}

\usepackage{bbding}

\def\BibTeX{{\rm B\kern-.05em{\sc i\kern-.025em b}\kern-.08em
    T\kern-.1667em\lower.7ex\hbox{E}\kern-.125emX}}
\begin{document}

\title{Privacy-Preserved Neural Graph Similarity Learning}

\author{
\IEEEauthorblockN{Yupeng Hou\textsuperscript{1,3}, Wayne Xin Zhao\textsuperscript{1,3,4 \Letter}, Yaliang Li\textsuperscript{2}, and Ji-Rong Wen\textsuperscript{1,3}}
\IEEEauthorblockA{\textsuperscript{1} \textit{Gaoling School of Artificial Intelligence, Renmin University of China}\\
\textit{\textsuperscript{2} Alibaba Group}\\
\textit{\textsuperscript{3} Beijing Key Laboratory of Big Data Management and Analysis Methods}\\
\textit{\textsuperscript{4} Engineering Research Center of Next-Generation Intelligent Search and Recommendation, Ministry of Education}\\
\{houyupeng,jrwen\}@ruc.edu.cn, batmanfly@gmail.com, yaliang.li@alibaba-inc.com}\thanks{\Letter\ Corresponding author.}
}

\newcommand{\ie}{\emph{i.e.,}\xspace}
\newcommand{\eg}{\emph{e.g.,}\xspace}
\newcommand{\aka}{\emph{a.k.a.,}\xspace}
\newcommand{\etal}{\emph{et al.}\xspace}
\newcommand{\paratitle}[1]{\vspace{1.5ex}\noindent\textbf{#1}}
\newcommand{\wrt}{w.r.t.\xspace}
\newcommand{\ignore}[1]{}

\newcommand{\tba}{\textcolor{red}{xxx }}
\newcommand{\ours}{PPGM }

\newtheorem{myDef}{Definition}

\maketitle

\begin{abstract}
To develop effective and efficient graph similarity learning (GSL) models, a series of data-driven neural algorithms have been proposed in recent years. Although GSL models are frequently deployed in privacy-sensitive scenarios, the user privacy protection of neural GSL models has not drawn much attention.
To comprehensively understand the privacy protection issues, we first introduce the concept of \emph{attackable representation} to systematically characterize the privacy attacks that each model can face.
Inspired by the qualitative results, we propose a novel \underline{P}rivacy-\underline{P}reserving neural \underline{G}raph \underline{M}atching network model, named \underline{PPGM}, for graph similarity learning. 
To prevent reconstruction attacks, the proposed model does not  communicate node-level representations between devices. Instead, we learn multi-perspective graph representations based on learnable context vectors. 
To  alleviate the attacks to graph properties, the obfuscated features that contain information from both graphs are communicated. In this way, the private properties of each graph can be difficult to infer. Based on the  node-graph matching techniques while calculating the obfuscated features, PPGM can also be effective in similarity measuring. To quantitatively evaluate the privacy-preserving ability of neural GSL models, we further propose an evaluation protocol via training supervised black-box attack models. Extensive experiments on widely-used benchmarks show the effectiveness and strong privacy-protection ability of the proposed model PPGM. The code is  available at: \textcolor{blue}{\url{https://github.com/RUCAIBox/PPGM}}.
\end{abstract}


\section{Introduction}\label{sec:intro}

Graph similarity learning (GSL) is one of the most fundamental tasks
in the literature of graph machine learning, intending to quantify the similarity of two given graphs~\cite{ma2021gsl_survey}. 
Various methods have been proposed to improve the performance of graph similarity learning methods, from early algorithms based on graph edit distance (GED)~\cite{bunke1997ged} or maximum common subgraph (MCS)~\cite{bunke1998mcs} metrics to the later graph kernels~\cite{niko2017matching,horv2004cyclic,yanardag2015deep_graph_kernels}.
However, these methods typically require exponential time complexity, largely limiting the application on real-world tasks.
To further improve the performance and efficiency of GSL models, a series of data-driven approximate approaches based on graph neural networks (GNNs) have been proposed recenrly~\cite{bai2019simgnn,li2019gmn,bai2020graphsim}. These methods greatly broaden the application scope of GSL to more realistic-sized graph data.

A basic setting of graph similarity learning assumed in most methods is  that the entire graph data is available to use from user devices, \ie nodes, edges and the corresponding features, while the assumption is usually not realistic in real scenarios.
We observe that GSL models are frequently used in privacy-sensitive scenarios, such as binary function similarity search~\cite{li2019gmn}, healthcare data management~\cite{niko2017matching}, and user portrait matching in recommender systems~\cite{su2021gmcf}. For example, when using code-checking systems based on GSL algorithms, users may upload their compiled programs (as binary function graphs~\cite{li2019gmn}) from user devices. 
These uploaded graph data with private information should be carefully protected, as they
are usually at high risk of privacy attacks. 
External attackers may intercept the uploaded graphs from the communication (\eg uploading) or disguise themselves as a fake data center to steal these graph data with user privacy. 
As a result, it is necessary to care about the privacy issue while developing graph similarity learning methods.

\begin{figure}[!t]
    \centering
    \includegraphics[width=0.95\linewidth]{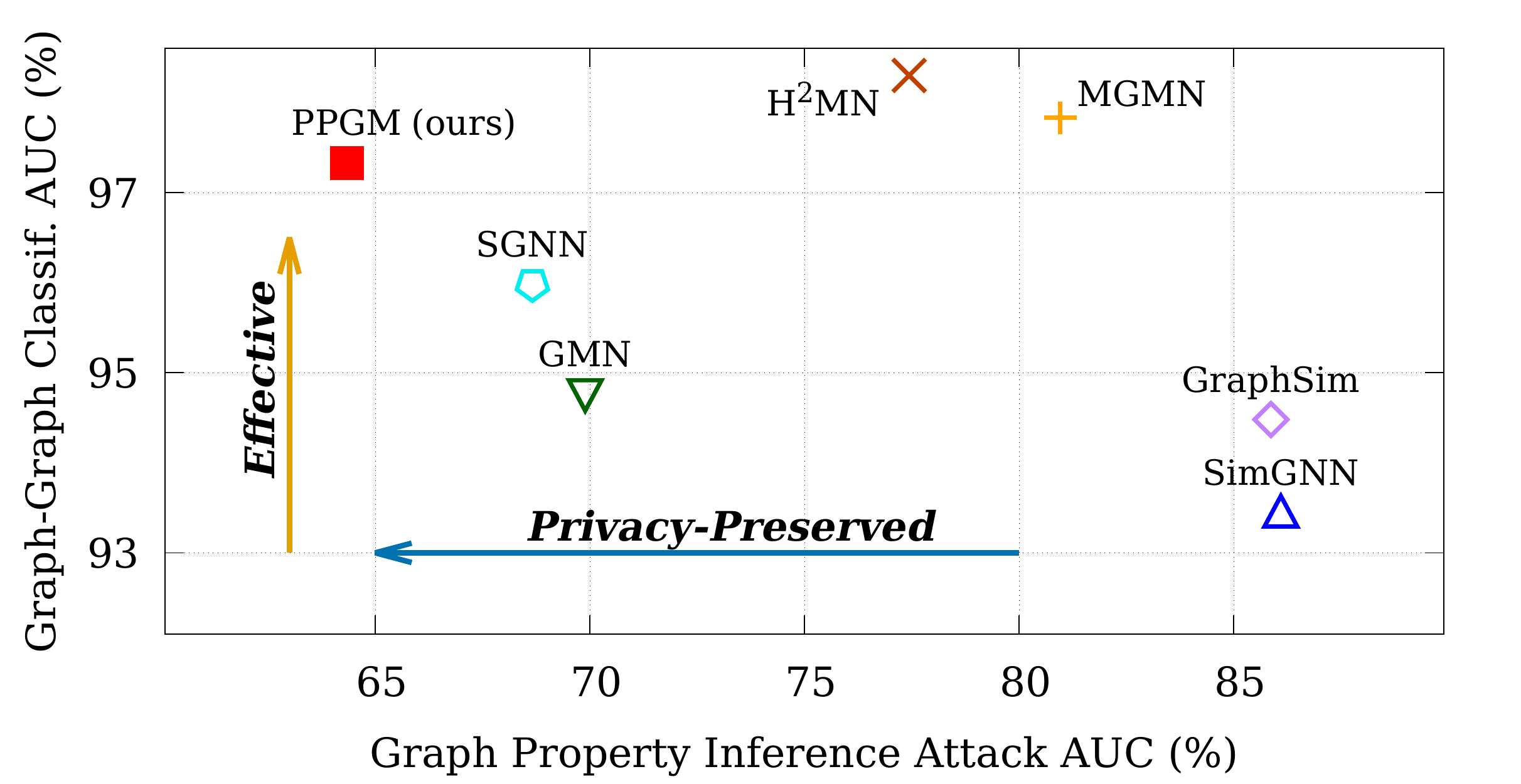}
  \caption{Performance comparison of different neural graph similarity learning models on privacy protection and graph-graph classification tasks on FFmpeg [50,200] dataset.}
    \label{fig:abs}
\end{figure}

However, it is challenging
to conduct both \emph{privacy-preserved} and \emph{effective} graph similarity learning models.
Although there have been several privacy protection techniques proposed for graph neural networks~\cite{ren2018lopub,wu2021fedgnn} or general multi-party computation approaches~\cite{craig2009homo},
these techniques usually can not be directly applied to the graph similarity learning tasks.
For example, graph publishing~\cite{zhao2016graph_pub,day2016publishing} and local differential privacy~\cite{ren2018lopub,cormode2018privacy} are widely studied, which  directly modify the raw graph data or learned representations and introduce randomness.
However, the modifications make it even harder to measure precise graph similarity scores. The introduced noises are almost certain to hurt the similarity measuring performance.
Besides, the concept of secure multiparty computation~\cite{yao1982protocols,yao1986generate} and homomorphic encryption~\cite{rivest1978data} may potentially benefit the GSL tasks, as the input graphs can be naturally seen as multiple parties that involve in the computation. However, existing solutions that fulfill the conceptions~\cite{craig2009homo} typically necessitate extreme computation and only support a few simple operators. Due to these restrictions, it's difficult to address complex non-linear neural networks with the above-mentioned algorithms.
As a result, we should  design special privacy-preserving graph similarity learning models, taking a comprehenisve consideration of the effectiveness, efficiency, and privacy-preservation trade-offs.

To tackle the above challenges, the basic idea of our approach is to deploy neural networks on user devices in a distributed computing environment~\cite{he2021fedgraphnn,wang2022fedscope_gnn}. 
We would like to keep most graph calculations on the user side.
In this way, once the device can be viewed as a trustworthy environment, then only the representations for communication between devices are at risk of privacy attacks.
Although several neural GSL models can be deployed in a distributed computing environment, we argue that the privacy-preserving abilities of these models still vary greatly.
The major reason is that these representations carry significantly different levels of user privacy information. Attackers can still leverage the representations sent off the devices to reconstruct graph structure or infer properties with user privacy.
As a result, it is necessary to analyze the privacy leakage level for each model. Then we can accordingly design highly privacy-preserved and effective GSL models, making the communicated representations hard to attack.

To this end, we first 
introduce the concept of \emph{attackable representations} to qualitatively analyze the types of potential privacy attacks when a neural GSL model is deployed in distributed computing environments.
We then propose a \textbf{P}rivacy-\textbf{P}reserved neural \textbf{G}raph \textbf{M}atching network for graph similarity learning, named \textbf{PPGM}. The key point is to learn obfuscated graph representations that are used for communication. 
First, as no node representations are involved in communication, the proposed method can naturally prevent reconstruction attacks.
Second, each obfuscated feature is fused from representations provided by both the input graphs. In this way, properties of one single graph are difficult to infer from the obfuscated features, alleviating the property inference attacks. In detail, our approach takes a pair of graphs as input and learns preliminary node representations inside each device. Based on shared context code vectors, multi-perspective graph representations are 
learned and communicated as messages. Then graph-node matching is performed on each device to generate obfuscated features. As the attackers will have an equal chance to intercept a graph representation (\ie message) or an obfuscated feature from the communicated representations of PPGM, so that we can achieve the goal of privacy protection. Meanwhile, the learning of obfuscated features introduces comprehensive node-graph representation interactions, which make the proposed model also effective on graph similarity learning tasks.

To evaluate the proposed model PPGM, we conduct extensive experiments on widely-used benchmark datasets. In addition, we propose a quantitative way to evaluate the privacy-preserving ability of neural GSL models. Shadow datasets extracted from original benchmarks and well-trained GSL models are leveraged to train supervised black-box attack models. Experimental results demonstrate that the proposed approach is privacy-preserving as well as effective. In summary, the main contributions are highlighted as follows:
\begin{itemize}
    \item To the best of our knowledge, we are the first to emphasize the privacy-preserving  concerns for neural graph similarity learning models. We introduce the concept of \emph{attackable representations}, which is a useful tool to systematically analyze the privacy attacks that neural GSL models may suffer from. (Section~\ref{sec:attack_tasks})
    \item We propose a privacy-preserved neural graph similarity learning model PPGM. With obfuscated features communicated between user devices, our model can be effective in similarity measuring while preventing reconstruction attacks and alleviating graph property inference attacks. (Section~\ref{sec:method})
    \item We propose a protocol to quantitatively evaluate the privacy-preserving ability of neural GSL models via training supervised black-box attack models on shadow datasets. (Section~\ref{sec:eval_protocol})
\end{itemize}

\section{Task}

In this section, we first briefly introduce the problem formulation and notation for graph similarity learning tasks. Then, we introduce the concept of \emph{attackable representations}, which is useful to qualitatively analyze the potential privacy attacks for neural GSL models.
Finally, we summarize three kinds of privacy attacks that neural GSL models may suffer from when deployed in a distributed computing environment.

\begin{figure}[t]
    \centering
    \includegraphics[width=0.39\textwidth]{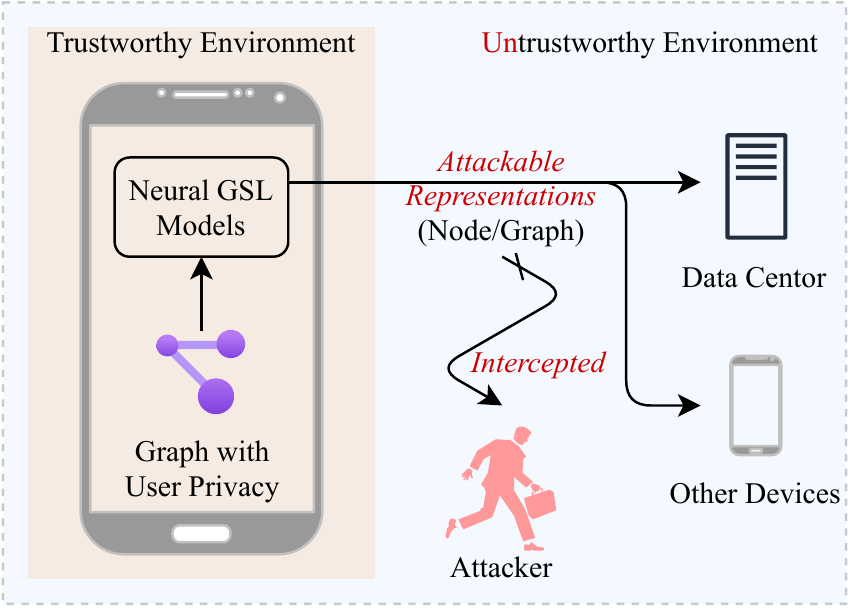}
    \caption{Illustration of the privacy attacks on \emph{attackable representations} while deploying neural GSL models in a distributed computing environment. In this work, we hypothesize that only the user devices are trustworthy environments, while any representations sent off the devices (\ie into untrustworthy environments) may be intercepted by attackers.}
    \label{fig:attack}
\end{figure}

\subsection{Graph Similarity Learning Tasks}

Given a pair of graph $\langle G_1,G_2\rangle$ and the corresponding label $y$, the task of graph similarity learning is to predict $y$ by mining the structure and attribute correlation between the given graphs. We have the first graph $G_1 = \{\mathcal{V}_1,\bm{X}_1, \mathcal{E}_1\}$, where $\mathcal{V}_1$ and $\mathcal{E}_1$ denotes the node set and edge set of $G_1$, respectively. $\bm{X}_1 \in \mathbb{R}^{|\mathcal{V}_1|\times f}$ denotes the node feature matrix, where $|\mathcal{V}_1|$ is the number of nodes in $G_1$ and $f$ is the feature dimension. The second graph $G_2 = \{\mathcal{V}_2,\bm{X}_2, \mathcal{E}_2\}$ can be formulated similarly. Based on whether labels are discrete or continuous, the graph similarity measuring tasks can be categorized into: (1) graph-graph classification, \ie $y\in \{0, 1\}$; (2) graph-graph regression, \ie $y \in [0, 1]$.

\begin{table}[t]
\caption{Attackable representations and potential privacy attacks of existing neural GSL models.}
\centering
\label{tab:task}
\begin{tabular}{@{}ccccc@{}}
\toprule
\multicolumn{1}{c}{Model} & \multicolumn{1}{c}{\begin{tabular}[c]{@{}c@{}}Attackable\\ Representations\end{tabular}} & \multicolumn{1}{c}{\begin{tabular}[c]{@{}c@{}}Reconstruction\\ Attack\end{tabular}} & \multicolumn{1}{c}{\begin{tabular}[c]{@{}c@{}}Graph Property\\ Inference\end{tabular}} \\ \midrule
SimGNN~\cite{bai2019simgnn}   & N, G & \textcolor{purple}{\CheckmarkBold} & \multirow{12}{*}{\begin{tabular}[c]{@{}c@{}}\textcolor{purple}{\CheckmarkBold}\\ \\ (All the neural\\ GSL models may\\ be threatened by\\ graph property\\ inference attacks)\end{tabular}} \\
GMN~\cite{li2019gmn}      & N, G & \textcolor{purple}{\CheckmarkBold} & \\
GraphSim~\cite{bai2020graphsim} & N    & \textcolor{purple}{\CheckmarkBold} & \\
SGNN~\cite{ling2021multilevel}     & G    & \textcolor{teal}{\XSolidBrush} & \\
MGMN~\cite{ling2021multilevel}     & N, G & \textcolor{purple}{\CheckmarkBold} & \\
H$^2$MN~\cite{zhang2021h2mn}  & N, H & \textcolor{purple}{\CheckmarkBold} & \\
EGSC-T~\cite{qin2021egsc}   & N    & \textcolor{purple}{\CheckmarkBold} & \\
EGSC-S~\cite{qin2021egsc}   & G    & \textcolor{teal}{\XSolidBrush} & \\\cmidrule(lr){1-3}
PPGM (ours)     & G, O & \textcolor{teal}{\XSolidBrush} & \\ \bottomrule
\multicolumn{4}{l}{\begin{tabular}[c]{@{}l@{}}``\textbf{N}'' - node representations, ``\textbf{G}'' - graph representations,\\ ``\textbf{H}'' - hypergraph representations, ``\textbf{O}'' - obfuscated features.\end{tabular}}
\end{tabular}
\end{table}

\subsection{Attackable Representations of Neural GSL Models}\label{sec:attackable_rep}

For privacy protection concerns, one can deploy neural GSL models on each user device in a distributed computing environment and communicate representations with other devices. However, attackers are still able to infer privacy properties from these representations. To study how privacy attacks may threaten neural GSL models, 
the first step is to define which representations can be attacked.

Thus, we introduce the concept of \emph{attackable representations} as a helpful tool to filter out which representations of a neural GSL model can be attacked.
Given a neural graph similarity learning model, the node and graph representations that are directly involved in the calculation (\eg operators of addition and multiplication) with representations of the other graphs are called attackable representations.
An illustration of attackable representations in the deployment of neural GSL models is presented in Figure~\ref{fig:attack}.
Note that the set of attackable representations is the least essential data that must be sent off the trustworthy devices. Because according to the above description, representations except attackable representations can all be calculated inside user devices and have no need to be sent for communication.

Using the introduced concept attackable representations, attacks on neural GSL models can be simplified as attacks on the set of attackable representations.
Qualitatively, one can summarize the types of potential privacy attacks for a neural GSL model given its set of attackable representations. For example, for models that do not communicate node representations, it will be nearly impossible for attackers to reconstruct the structure of input graphs. Quantitatively, we can further benchmark the privacy-preserving abilities of neural GSL models. A possible way is to train attack models with attackable representations as input, and evaluate the attack performances as metrics on privacy-preserving ability.

\subsection{Privacy Attacks on Neural GSL Models}\label{sec:attack_tasks}

A common design for privacy attacks is supervised black-box privacy attacks~\cite{wang2021privacy}. The attack models are trained with both the attacker-prepared shadow dataset and the attackable representations, \eg usually intercepted from the deployed API services.
In what follows, we describe several privacy attack tasks against graph similarity learning models. Table~\ref{tab:task} systematically summarizes the attackable representations of several existing neural GSL models and whether they will suffer from specific privacy attacks described below.

\subsubsection{Structure Reconstruction Attack}\label{sec:structure-reconstruct}

Recently proposed GSL models usually extract features from node-node matching score matrices for accurate graph matching, such as SimGNN~\cite{bai2019simgnn}, GMN~\cite{li2019gmn}, and GraphSim~\cite{bai2020graphsim}.
The node representations of these models will unavoidably be sent from devices, serving as attackable representations.
With link prediction optimization objective, the topological structure of the input graphs may be reconstructed from these attackable node representations~\cite{wang2021privacy}.
For models that do not have node representations in the attackable representation set, \eg SGNN~\cite{ling2021multilevel}, the structure reconstruction attack can be prevented naturally.

\subsubsection{Attribute Reconstruction Attack}\label{sec:attr-reconstruct}

Besides topological structures of input graphs, the node attributes with user privacy may also be reconstructed by supervised privacy attack models. Generally, the node representations of the target samples are required to be attackable representations. As a result, those neural GSL models that can be attacked by structure reconstruction tasks may also be threatened by attribute reconstruction attacks.
Both the \emph{structure} and \emph{attribute} reconstruction attack tasks are referred to as \emph{reconstruction attack tasks} in the following sections.

\begin{figure*}[ht]
    \centering
    \includegraphics[width=0.9\textwidth]{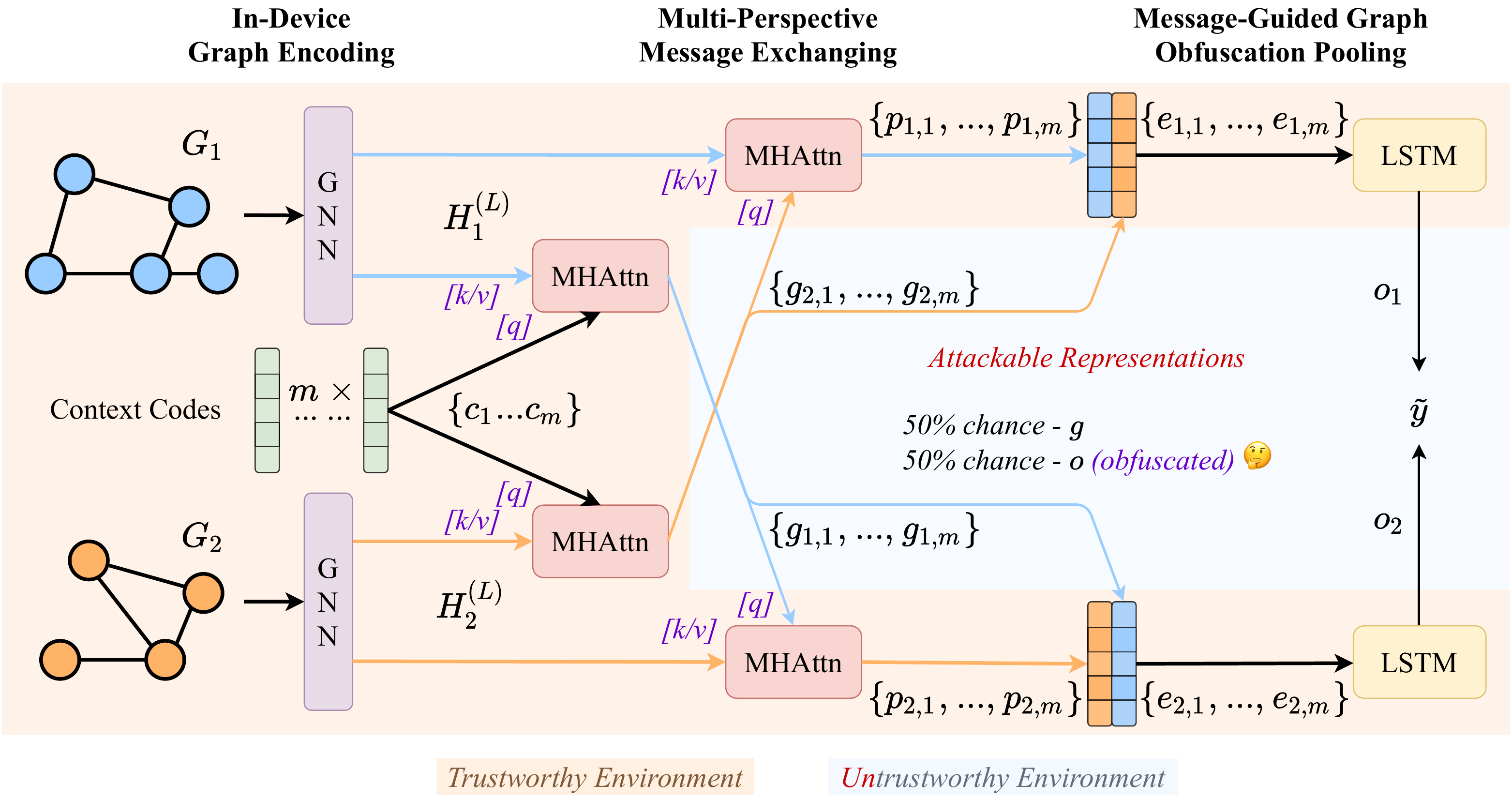}
    \caption{The overall framework of our proposed privacy-preserved neural graph similarity learning model PPGM.}
    \label{fig:model}
\end{figure*}

\subsubsection{Graph Property Inference Attack}\label{sec:prop-inf}

In addition to node-level and edge-level attack tasks, graph property inference attack tasks also threaten existing neural GSL models. Attackers usually prepare shadow datasets with labeled graph properties and train graph classification models to infer the properties with user privacy. Unlike structure and attribute reconstruction attacks, the graph property inference attack can threaten all the models. That is because once communication exists, we can always try to infer target properties from the intercepted data, no matter what kinds of attackable representations are. Although unavoidable, as shown in our experiments in Section~\ref{sec:overall_performance}, different models may have varying privacy-preserving abilities against property inference attacks. Thus, it is also valuable to design neural GSL models for alleviating property inference attacks. We propose a protocol to quantitatively evaluate the privacy-preserving ability of neural GSL models via empirical black-box attacking experiments in Section~\ref{sec:eval_protocol}.

\section{Methodology}\label{sec:method}

In this section, we present the proposed \textbf{P}rivacy-\textbf{P}reserved \textbf{G}raph \textbf{M}atching network for graph similarity learning, named \textbf{PPGM}.
The proposed approach aims to evaluate the similarity of the provided two graphs,
meanwhile preventing reconstruction attacks (described in Section~\ref{sec:structure-reconstruct} and~\ref{sec:attr-reconstruct}) and mitigating graph property inference attacks (described in Section~\ref{sec:prop-inf}) when the well-trained model is deployed in a distributed computing environment.

We first introduce the graph encoding module inside user devices in Section~\ref{sec:in-device-gnn}, which outputs the preliminary node representations of each graph.
Then, in Section~\ref{sec:message}, we introduce how to extract and exchange graph representations from multiple perspectives as messages communicated between devices, which help match between graphs and prevent exposing node representations.
We further propose a graph obfuscation pooling layer in Section~\ref{sec:obfus-pool}, which performs node-graph matching to fuse the node representations of the current graph with messages revived from the other graph.
The obfuscated features will contribute to the graph similarity prediction, while the properties of every single graph are difficult to infer.
Finally, we make a brief discussion in Section~\ref{sec:discuss}

\subsection{In-Device Graph Encoding}\label{sec:in-device-gnn}

The proposed approach is generally deployed in a distributed computing environment. 
Most calculations are kept inside the user devices and only necessary representations should be sent off the devices. These representations are defined as \emph{attackable representations} in Section~\ref{sec:attack_tasks}.
Thus, we firstly encode graph structure and attributes into preliminary node representations via the graph encoding layer inside user devices.

Recent years have witnessed the rapid growth of graph neural networks (GNNs)~\cite{zhou2020gnn_survey}, such as GCN~\cite{kipf2017gcn}, GAT~\cite{velickovic2018gat}, GraphSAGE~\cite{hamilton2017graphsage}, and GIN~\cite{xu2019gin}. By recursive aggregating neighbor representations of each node, GNNs can extract effective node and graph representations. As a result, we adopt stacked GNN layers as in-device graph encoders here. Formally, take GCN~\cite{kipf2017gcn} as example, for an input graph $G_1 = \{\mathcal{V}_1, \bm{X}_1, \mathcal{E}_1\}$, we have:
\begin{align}
    \bm{H}^{(l)}_1 = \sigma \left(\tilde{\bm{A}}_1\bm{H}^{(l-1)}_1\bm{W}^{l}\right),\label{eq:gnn}
\end{align}
where $\tilde{\bm{A}}_1 \in \mathbb{R}^{|\mathcal{V}_1|\times |\mathcal{V}_1|}$ denotes the normalized Laplacian matrix of graph $G_1$, $\bm{H}^{(l)}_1 \in \mathbb{R}^{|\mathcal{V}_1|\times d}$ denotes the hidden states of GCN at layer $l$, $\bm{W}^l \in \mathbb{R}^{d\times d}$ denotes the learnable parameters at layer $l \ge 1$, and $d$ is the dimension of output node representations. Note that $\bm{H}_1^{(0)} = \bm{X}_1 \in \mathbb{R}^{|\mathcal{V}_1|\times f}$ is the node  attributes and $\bm{W}^0 \in \mathbb{R}^{f\times d}$ are learnable parameters at layer $l = 0$.
We can obtain preliminary node representations from the output of the in-device graph encoder's last layer, \ie $\bm{H}_1^{(L)}$ for $G_1$ and $\bm{H}_2^{(L)}$ for $G_2$, where $L$ is the total number of GNN layers. It is also possible to replace GCN with more expressive GNNs like GIN~\cite{xu2019gin,zhang2021h2mn} in the proposed framework. 

\subsection{Multi-Perspective Message Exchanging}\label{sec:message}

After obtaining preliminary node representations of each graph, we extract graph representations from multiple perspectives as messages and communicate them among devices for further matching. 

\subsubsection{Context-Attentive Message Extraction Layer}

Although node representations on each device have been learned via graph neural networks, directly sending node representations for further matching will put the user privacy at risk. As discussed in Section~\ref{sec:structure-reconstruct} and Section~\ref{sec:attr-reconstruct}, node representations are attackable for both structure and attribute reconstruction attacks.
The node representations contain rich semantics of both the centor node and the $L$-hop context of the ego-graph. The neighbor node context has shown to be helpful at predicting the attributes of the centor node~\cite{hu2019strategies}, further risking the node-level privacy.

Considering the above issues, we choose to send graph representations instead of node representations off the devices, so that we can perform node-graph matching on each device later as well as prevent reconstruction attacks.
Generally, we can pool node representations as graph representations via heuristic methods, \eg max pooling, from a single perspective. 
However, such heuristic pooling techniques are naturally unsupervised graph representation learning methods~\cite{xu2019gin}. The pooled graph representations may not be effective for graph similarity learning tasks, instead, the properties with user privacy can be easily inferred. As compared, the supervised attention-based graph pooling methods may learn to extract features that are useful for similarity learning and abandon those features containing private graph properties.
Besides, intuitively, graph representations from multiple perspectives may be helpful for more comprehensive matching with the other graphs than those that are just extracted from a single perspective.
Thus, we propose to extract multi-perspective graph representations via a multi-head attention-based graph pooling module.

Detailed, to pool a graph from multiple perspectives, inspired by recent advances on passage encoding~\cite{humeau2020poly}, we propose to learn $m$ context codes $\{\bm{c}_1, \ldots, \bm{c}_m\}$, where each context code is a learnable vector, \ie $\bm{c}_i \in \mathbb{R}^{d}, 1 \le i \le m$. Given a specific context code $\bm{c}_i$, we adopt the multi-head attention architecture~\cite{vaswani2017attention} to pool the node representations. We specially design \emph{queries} (context code vectors), \emph{keys} (node representations) and \emph{values} (node representations) as:
\begin{align}
    \bm{g}_{2,i} = \operatorname{MHAttn}(\bm{c}_i, \bm{H}_2^{(L)}, \bm{H}_2^{(L)}),\label{eq:g}
\end{align}
where $\bm{H}_2^{(L)}$ denotes the output node representations of in-device graph encoders, and $\bm{g}_{2,i}$ is the pooled graph representations of $G_2$ under context $\bm{c}_i$. Those node representations (as values) that are more similar to the context code (key-query similarity) receive larger attention weights in the final graph representations. Note that different devices share a common series of well-trained context codes while deploying. The context codes are trained to extract features that are useful for similarity measuring. The pooled representations $\{\bm{g}_{1,1},\ldots,\bm{g}_{1,m}\}$ and $\{\bm{g}_{2,1},\ldots,\bm{g}_{2,m}\}$ are attentive to the same context code at each position $1\le i \le m$.

\subsubsection{Cross-device Message Communication}

So far, the encoding of $G_1$ and $G_2$ are all in-device and there is no communication between devices. After obtaining multi-perspective graph representations, these representations serve as messages and be sent to the other devices for further matching.

As no node-level representations are included as messages, the communication process naturally prevent reconstruction attacks. 
Although exchanging graph-level representations may be threatened by graph property inference attacks, it's unavoidable for neural GSL methods.
To measure the similarity of two graphs, their original graph data or encoded representations finally fall into one of the devices, either a user device or a centralized data center. As a result, at least one of the user devices will send messages, which make private graph properties at risk and attackable.

Although the messages alone may be at risk of privacy leakage, we point out that the messages together with the final obfuscated representations (which will be introduced next) can make property inference attacks difficult.

\subsection{Message-Guided Graph Obfuscation Pooling}\label{sec:obfus-pool}

Given node representations (from the current device) and graph representations (carried by messages from the other devices), we fuse these representations as obfuscated features for further prediction and meanwhile alleviating the property inference attacks.

\subsubsection{Message-Attentive Graph Pooling Layer}

Without loss of generality, we take $G_1$ as an example here.
After communication among devices, we have node representations $\bm{H}_1^L$ (Eqn.~\eqref{eq:gnn}) and graph representations $\{\bm{g}_{2,1},\ldots,\bm{g}_{2,m}\}$ (Eqn.~\eqref{eq:g}) from the other device (\eg from device of $G_2$).

We then perform node-graph matching for better similarity learning. Here we utilize another multi-head attention layer and adopt a message-guided graph pooling method. For each message $\bm{g}_{2,i}$, we calculate the attention weight for each node representation (as \emph{keys} and \emph{values}) according to its similarity to the message (as \emph{queries}). In this way, the node-graph matching is performed naturally by calculation of attention weights, as:
\begin{equation}
    \bm{p}_{1,i} = \operatorname{MHAttn}(\bm{g}_{2,i}, \bm{H}_{1}^{(L)}, \bm{H}_1^{(L)}),\label{eq:p}
\end{equation}
where $\bm{p}_{1,i} \in \mathbb{R}^{d}$ is the pooled representation of $G_1$ guided by the message $\bm{g}_{2,i}$. Here on the device of $G_1$, we have the message-attentive graph representations $\{\bm{p}_{1,1}, \ldots,\bm{p}_{1,m}\}$ and the context-attentive graph representations $\{\bm{g}_{2,1},\ldots,\bm{g}_{2,m}\}$ (sent from the device of $G_2$). These representations are corresponding at each position.

\subsubsection{Ordered Global Representation Obfuscation}

Once we have graph representations of both graphs on each device, we then fuse the representations together into obfuscated features. The obfuscated features will be sent off the device for final prediction.
Without loss of generality, we can define an obfuscation function $f_{\operatorname{OBF}}(\cdot): \mathbb{R}^{2m\times d}\rightarrow\mathbb{R}^{d}$ as:
\begin{align}
    \bm{o}_1 = f_{\operatorname{OBF}}\left(\{\bm{p}_{1,1}, \ldots,\bm{p}_{1,m}\}, \{\bm{g}_{2,1},\ldots,\bm{g}_{2,m}\}\right),
\end{align}
where $\bm{o}_1$ is the obfuscated feature of graph $G_1$.

Ideal obfuscated features should have two characteristics: (a) Be able to indicate the similarities between graphs well, \eg similar graphs have similar obfuscated features; (b) Contain information from both graphs, which means that the properties of each graph are hard to infer.
Under such instructions, we propose one implementation of the obfuscation function $f_{\operatorname{OBF}}(\cdot)$. We first concatenate message-attentive and context-attentive graph representations at each position in order:
\begin{align}
    \bm{e}_{1,i} = \bm{p}_{1,i}||\bm{g}_{2,i},\label{eq:e}
\end{align}
where $\bm{e}_{1,i} \in \mathbb{R}^{2d}$ is the concatenated representation in position $i$ of graph $G_1$, and ``$||$'' denotes the concatenation operation. We then place the concatenated representations in order, and apply a sequence model for further fusion and obfuscation:
\begin{align}
    \bm{o}_1 = \operatorname{LSTM}(\bm{e}_{1,1}, \ldots,\bm{e}_{1,m}),
\end{align}
where $\operatorname{LSTM}$ is the long short-term memory network~\cite{hochreiter1997lstm}. The proposed design is able to have the above characteristics: (a) Two similar graphs have similar ordered obfuscated features, as they have similar $\bm{e}_i$ on each position, which is helpful for similarity measuring;
(b) The graph representations from the two graphs are fused by LSTM, making graph properties hard to infer from a single obfuscated feature.
Note that the LSTM network here is introduced for fusing the representations from both graphs, which could also be replaced by other functions and we leave it as our future work.

\subsubsection{Training and Prediction}

Finally, the obfuscated features $\bm{o}_1$ and $\bm{o}_2$ are sent off the devices for matching score calculation.
For graph-graph classification task, we can measure the graph similarity based on their cosine similarity in the representation space as $\tilde{y}_c = \operatorname{cosine}(\bm{o}_1, \bm{o}_2)$.
For graph-graph regression task, we can measure the similarity score based on multilayer perceptron (MLP) layers as $\tilde{y}_r = \sigma(\operatorname{MLP}([\bm{o}_1;\bm{o}_2]))$,
where $\sigma(\cdot)$ is the sigmoid function. While training, we can optimize the parameters with mean squared error (MSE) loss function.

\subsection{Discussion}\label{sec:discuss}

Overall, the proposed approach can prevent reconstruction attacks and alleviate property inference attacks. The attackable representations for $G_1$ are $\bm{g}_{1,i}$ (Eqn.~\eqref{eq:g}) and $\bm{p}_{1,i}$ (Eqn.~\eqref{eq:p}). Due to the carefully designed architecture of PPGM, there are no node representations in the attackable representation set, preventing the reconstruction attacks naturally. Besides, for attackers, the representations intercepted from the deployed API services have an equal chance to be $\bm{g}_{1,i}$ or $\bm{p}_{1,i}$. As nearly half of the intercepted representations are obfuscated, the  property inference attacks can be greatly alleviated. 

Note that PPGM focuses on privacy protection after being deployed in distributed computing environments, but not during the model training (\eg usually achieved by federated learning~\cite{wu2021fedgnn,xie2021federated,he2021fedgraphnn,wang2022fedscope_gnn}).
However, in real scenarios of graph similarity learning, the privacy issues for deploying are more important than those for training.
Most similarity labels in the training set are generated directly from raw data of the given pairs of graphs, making it unrealistic to consider privacy protection while training neural GSL models.

It is also worth noting that PPGM is highly parameter-efficient, since both GNNs and multi-head attention layers do not require very deep neural networks. The overall space complexity of PPGM is $O\left((|\mathcal{V}| + d)^2L + m\cdot d\right)$. For a typical hyperparameter setting with $d=100, L=3$ and $m=8$, the model usually consumes around $1$ \textsc{Mb} memories. As a result, the well-trained PPGM models are suitable to be installed or deployed in portable devices, keeping most calculations inside user devices.

\begin{table}[t]
\caption{Statistics of the datasets. ``\#$|\mathcal{G}|$'' denotes the number of graphs. ``Avg. $|\mathcal{V}|$'' denotes the average number of nodes. ``Avg. $|\mathcal{E}|$'' denotes the average number of edges. ``Target Properties'' denotes the labels used for property inference experiments.}
\label{tab:dataset}
\centering
\begin{tabular}{@{}crrr@{}}
\toprule
Datasets             & \multicolumn{1}{c}{\#$|\mathcal{G}|$} & \multicolumn{1}{c}{Avg. $|\mathcal{V}|$} & \multicolumn{1}{c}{Avg. $|\mathcal{E}|$}                       \\ \midrule
FFmpeg {[}20,200{]}  & 31,696                              & 51.02                                    & 75.88                                    \\
FFmpeg {[}50,200{]}  & 10,824                              & 90.93                                    & 136.83                                                                                                                              \\
OpenSSL {[}20,200{]} & 15,800                              & 44.89                                    & 67.15                                                                                                                               \\
OpenSSL {[}50,200{]} & 4,308                               & 83.68                                    & 127.75                                                                                                                             \\
\bottomrule
\end{tabular}
\end{table}

\begin{table*}[!t]
\caption{Performance comparison on the graph-graph classification task and the corresponding property inference attack in terms of AUC scores (\%). ``Classif.'' denotes ``Classification''. Higher $\uparrow$ is better for classification metrics, while lower $\downarrow$ is better for attack tasks. We highlight the top-3 best results in each column and label their ranks.}
\label{tab:classification}
\centering
\begin{tabular}{@{}cllllllll@{}}
\toprule
Dataset    & \multicolumn{2}{c}{FFmpeg {[}20, 200{]}} & \multicolumn{2}{c}{FFmpeg {[}50, 200{]}}   & \multicolumn{2}{c}{OpenSSL {[}20, 200{]}} & \multicolumn{2}{c}{OpenSSL {[}50, 200{]}} \\
\cmidrule(lr){2-3}\cmidrule(lr){4-5}\cmidrule(lr){6-7}\cmidrule(lr){8-9}
Task       & \multicolumn{1}{c}{Attack $\downarrow$} & \multicolumn{1}{c}{Classif. $\uparrow$}   & \multicolumn{1}{c}{Attack $\downarrow$}       & \multicolumn{1}{c}{Classif. $\uparrow$}   & \multicolumn{1}{c}{Attack $\downarrow$}  & \multicolumn{1}{c}{Classif.    $\uparrow$}      & \multicolumn{1}{c}{Attack $\downarrow$}         & \multicolumn{1}{c}{Classif.  $\uparrow$}        \\ \midrule\midrule
SimGNN     & 86.05\textsubscript{$\pm$0.49} & 94.31\textsubscript{$\pm$1.01} & 86.10\textsubscript{$\pm$0.73} & 93.45\textsubscript{$\pm$0.54} & 91.34\textsubscript{$\pm$0.32} & 93.58\textsubscript{$\pm$0.82} & 82.90\textsubscript{$\pm$0.74} & 94.25\textsubscript{$\pm$0.85} \\
GMN        & 78.50\textsubscript{$\pm$1.59} & 95.92\textsubscript{$\pm$1.38} & 74.02\textsubscript{$\pm$0.78} & 94.76\textsubscript{$\pm$0.45} & 90.74\textsubscript{$\pm$0.10} & 93.03\textsubscript{$\pm$3.81} & 82.69\textsubscript{$\pm$1.25} & 93.91\textsubscript{$\pm$1.65} \\
GraphSim   & 85.46\textsubscript{$\pm$2.03} & 96.49\textsubscript{$\pm$0.28} & 85.87\textsubscript{$\pm$0.59} & 94.48\textsubscript{$\pm$0.73} & 92.34\textsubscript{$\pm$0.26} & 94.97\textsubscript{$\pm$0.98} &  87.52\textsubscript{$\pm$0.31} & 93.66\textsubscript{$\pm$1.84} \\
SGNN       & \textbf{70.70\textsubscript{$\pm$0.59} (3)} & 96.29\textsubscript{$\pm$0.14} & \textbf{68.66\textsubscript{$\pm$1.13} (3)} & 95.98\textsubscript{$\pm$0.32} & \textbf{74.82\textsubscript{$\pm$1.13} (2)} & 93.79\textsubscript{$\pm$0.17} & \textbf{63.74\textsubscript{$\pm$1.09} (2)}  & 93.21\textsubscript{$\pm$0.82} \\
MGMN       & 83.46\textsubscript{$\pm$0.62} & \textbf{98.29\textsubscript{$\pm$0.10} (2)} & 80.96\textsubscript{$\pm$1.03} & \textbf{97.83\textsubscript{$\pm$0.11} (2)} & 90.01\textsubscript{$\pm$0.42} & \textbf{97.59\textsubscript{$\pm$0.24} (2)} & 74.94\textsubscript{$\pm$2.55} & \textbf{95.58\textsubscript{$\pm$1.13} (2)} \\
H$^2$MN    & 79.27\textsubscript{$\pm$0.30} & \textbf{98.54\textsubscript{$\pm$0.14} (1)} & 77.44\textsubscript{$\pm$0.76} & \textbf{98.30\textsubscript{$\pm$0.29} (1)} & 90.69\textsubscript{$\pm$0.04} & \textbf{98.47\textsubscript{$\pm$0.38} (1)} & 87.36\textsubscript{$\pm$0.10} & \textbf{96.80\textsubscript{$\pm$0.95} (1)} \\
SGNN$_\text{LDP}$ 
           & \textbf{69.72\textsubscript{$\pm$0.98} (2)} & 95.87\textsubscript{$\pm$0.09} & \textbf{67.63\textsubscript{$\pm$0.64} (2)} & 95.22\textsubscript{$\pm$0.09} & \textbf{75.82\textsubscript{$\pm$0.96} (3)} & 93.68\textsubscript{$\pm$0.24} & \textbf{65.55\textsubscript{$\pm$1.69} (3)} & 92.49\textsubscript{$\pm$0.31} \\ \midrule
PPGM       & \textbf{67.88\textsubscript{$\pm$0.52} (1)} & \textbf{97.90\textsubscript{$\pm$0.07} (3)} & \textbf{64.34\textsubscript{$\pm$2.05} (1)} & \textbf{97.33\textsubscript{$\pm$0.15} (3)} & \textbf{63.35\textsubscript{$\pm$1.93} (1)} & \textbf{97.11\textsubscript{$\pm$0.29} (3)} & \textbf{56.05\textsubscript{$\pm$2.36} (1)} & \textbf{94.55\textsubscript{$\pm$0.61} (3)} \\ \bottomrule
\end{tabular}
\end{table*}

\section{Experiments}

To verify the effectiveness of the proposed \ours on privacy-preserving tasks as well as graph similarity learning tasks, we conduct extensive experiments and provide detailed result analysis.

\subsection{Privacy-Preserving Ability Evaluation}\label{sec:eval_protocol}

To quantitatively evaluate the privacy-preserving ability of neural GSL models, we propose an evaluation protocol based on training supervised black-box attack models. Note that as we have discussed in Section~\ref{sec:attack_tasks}, whether a model can be attacked by the reconstruction attacks can be analyzed qualitatively via the tool of \emph{attackable representations} introduced in Section~\ref{sec:attackable_rep}. As summarized in Table~\ref{tab:task}, several methods can prevent  reconstruction attacks (\eg the proposed model PPGM), while all the models may suffer from  graph property inference attacks. Thus, the proposed quantitative privacy-preserving  ability evaluation focuses on the property inference attack tasks and does not include reconstruction attack tasks.

We select the graph properties to attack from the original datasets, \ie the compiler (\eg gcc) and optimization level (\eg O3) for binary code similarity datasets.
We sample $10\%$ graphs from each dataset's original training set as shadow datasets for training black-box attack models~\cite{dai2022comprehensive} and directly use all the graphs in the original validation and test sets for evaluation.
We encode each graph in the shadow dataset with well-trained neural GSL models. The attackable representations are saved for training attack models.

We adopt another randomly initialized multilayer perceptron (MLP) as the attack model. As for the input of attack models (\ie attackable representations), the node representations are pooled and concatenated with graph representations to simulate a random interception attack.

\subsection{Experimental Setup}

\subsubsection{Datasets}

To evaluate the performance of the proposed PPGM, we conduct experiments on public benchmark datasets.
In detail, we evaluate models on two binary code similarity datasets \textbf{FFmpeg}\footnote{https://www.ffmpeg.org/} and \textbf{OpenSSL}\footnote{https://www.openssl.org/}~\cite{xu2017binary,ling2021multilevel}. Following previous work~\cite{ling2021multilevel}, we use two subsets $[20,200]$ and $[50,200]$ that limit the minimum and the maximum number of nodes for evaluating performance on graphs at different scale levels.
We use the same train/validation/test splits following previous works~\cite{ling2021multilevel,zhang2021h2mn}.
The statistics of the datasets are summarized in Table~\ref{tab:dataset}.

\subsubsection{Compared Methods}

We compare the proposed method with the following baseline methods.

$\bullet$ \textbf{SimGNN}~\cite{bai2019simgnn} extracts histogram features from node-node matching score matrix for prediction.

$\bullet$ \textbf{GMN}~\cite{li2019gmn} introduces both intra-graph and inter-graph message passing techniques as graph matching networks.

$\bullet$ \textbf{GraphSim}~\cite{bai2020graphsim} proposes to extract features from node-node matching score matrix via hierarchical CNNs.

$\bullet$ \textbf{SGNN}~\cite{ling2021multilevel} leverages siamese graph neural networks and compares the graph representations.

$\bullet$ \textbf{MGMN}~\cite{ling2021multilevel} proposes to extract multi-level graph and node features for comprehensive matching.

$\bullet$ \textbf{H$^2$MN}~\cite{zhang2021h2mn} utilizes hypergraph pooling and convolution for efficient and accurate similarity measuring.

$\bullet$ \textbf{EGSC-T}~\cite{qin2021egsc} utilizes attention techniques for early-fusion of node-level representations.

$\bullet$ \textbf{EGSC-S}~\cite{qin2021egsc} has similar siamese architecture as SGNN. The model is distilled by EGSC-T.

$\bullet$ \textbf{SGNN$_\text{LDP}$} is a variant of SGNN to show the effectiveness of conventional privacy protection methods. We apply local differential privacy (LDP) techniques~\cite{ren2018lopub} on the graph representations. Detailed, in our implementation, we add zero-mean Laplacian noise on graph representations before communicating between two graphs.

\subsubsection{Evaluation Metrics}

Following previous works~\cite{ling2021multilevel,qin2021egsc}, we calculate the area under the ROC curve (AUC) for the graph-graph classification tasks.
For the property inference attack on graph-graph classification tasks, we also adopt AUC scores to evaluate the attack models. However, different from classification tasks where \emph{higher} AUC is better, in the property inference attack task, \emph{lower} AUC scores mean that the target neural GSL models have stronger privacy protection abilities.

\subsubsection{Implementation Details}

The proposed model \ours and all the baselines are implemented using PyTorch\footnote{https://pytorch.org/} 
and PyG~\cite{fey2019pyg}. We optimize all the methods with Adam optimizer~\cite{kingma2015adam} and search the hyperparameters for a fair comparison. The dimensions of hidden states are set to $d=100$. For the graph-graph classification task, we train the models for $100$ epochs with $5e^{-4}$ learning rate and $10$ pairs in each batch. 
For the property inference task, we train a $3$-layer MLP model with ReLU activation.
For hyperparameters of the proposed model PPGM, we tune the number of context codes $m \in \{4, 8, 16\}$ and use $4$ heads for the multi-head attention modules.
The results of baseline methods (except SGNN$_{\text{LDP}}$) on graph similarity learning tasks are inherited directly from the original papers.
For other results, we report metrics on the test set with models that gain the highest performance on the validation set.

\subsection{Overall Performance}\label{sec:overall_performance}

We compare the proposed approach with the baseline methods on the two graph-graph classification datasets (two subsets for each dataset).
The results are reported in Table~\ref{tab:classification}.

For the baseline methods, we can observe that models that leverage node-node matching score matrices as features (\ie SimGNN, GMN, GraphSim, MGMN, and H$^2$MN) generally have better performance than SGNN on graph similarity learning metrics, showing the effectiveness of modeling detailed node-level correlations between graphs. However, SGNN significantly outperforms these graph matching-based methods on privacy attack tasks, as the communicated node representations contain user privacy information and are weak for defending inference attacks. For SGNN$_\text{LDP}$, a baseline to verify the conventional privacy-preserving techniques, we observe that it generally has worse similarity measuring performance and better privacy-preserving performance than SGNN. The results show that adding random noise can indeed improve the privacy-preserving ability but leads to a performance drop, which is not an ideal solution.

Finally, by comparing the proposed approach PPGM with all the baselines, we can find that PPGM achieves the best privacy-preserving performance and competitive similarity measuring performance. Different from these baselines, the specially designed neural GSL model PPGM contains both obfuscated features and graph representations as attackable representations, making inferring properties of a single graph more difficult. Moreover, as PPGM introduces several techniques in consideration of the effectiveness, including
the multi-perspective messages and node-graph matching techniques, PPGM also has competitive similarity measuring performance. Note that besides the property inference attack, as shown in Table~\ref{tab:task}, the proposed model can prevent reconstruction attack either, further showing the privacy-preserving ability of PPGM. Overall, the results show that PPGM is highly privacy-preserved and effective.

\begin{table}[t]
\caption{Ablation study for the proposed model PPGM. The best performance is denoted in bold fonts.}
\label{tab:ablation}
\centering
\resizebox{\columnwidth}{!}{
\begin{tabular}{@{}lllll@{}}
\toprule
Dataset    & \multicolumn{2}{c}{FFmpeg {[}50, 200{]}}   & \multicolumn{2}{c}{OpenSSL {[}50, 200{]}} \\
\cmidrule(lr){2-3}\cmidrule(lr){4-5}
Task       & \multicolumn{1}{c}{Attack $\downarrow$} & \multicolumn{1}{c}{Classif. $\uparrow$}   & \multicolumn{1}{c}{Attack $\downarrow$}       & \multicolumn{1}{c}{Classif. $\uparrow$}  \\ \midrule\midrule
PPGM & \textbf{64.34\textsubscript{$\pm$2.05}} & \textbf{97.33\textsubscript{$\pm$0.15}} & \textbf{56.05\textsubscript{$\pm$2.36}} & \textbf{94.55\textsubscript{$\pm$0.61}} \\
\ \ $w/o$ Obfuscation & 66.47\textsubscript{$\pm$0.62} & 96.84\textsubscript{$\pm$0.25} & 60.97\textsubscript{$\pm$3.09} & 92.59\textsubscript{$\pm$0.90} \\
\ \ $w/o$ Context Codes & 86.23\textsubscript{$\pm$0.43} & 97.13\textsubscript{$\pm$0.16} & 88.14\textsubscript{$\pm$0.39} & 93.60\textsubscript{$\pm$0.80} \\
\ \ $w/o$ N-G Matching & 68.96\textsubscript{$\pm$1.62} & 97.06\textsubscript{$\pm$0.50} & 63.81\textsubscript{$\pm$1.90} & 93.90\textsubscript{$\pm$0.51} \\
SGNN       & 68.66\textsubscript{$\pm$1.13} & 95.98\textsubscript{$\pm$0.32} & 63.74\textsubscript{$\pm$1.09} & 93.21\textsubscript{$\pm$0.82} \\
\bottomrule
\end{tabular}
}
\end{table}

\subsection{Ablation Study}

In this part, we analyze how each of the proposed techniques or components affects the final performance. We prepare three variants of the proposed PPGM model for comparisons, including:

$\bullet$ \underline{$w/o$ Obfuscation} without feature obfuscation (\ie replace $\bm{g}_{2,i}$ to $\bm{g}_{1,i}$ in Eqn.~\eqref{eq:e}); 

$\bullet$ \underline{$w/o$ Context Codes} that replaces the context-attentive message extraction layer in Eqn.~\eqref{eq:g} with mean pooling;

$\bullet$ \underline{$w/o$ N-G Matching} that replaces the message-attentive graph pooling layer in Eqn.~\eqref{eq:p} with mean pooling.

The experimental results of the proposed model PPGM and its variants are reported in Table~\ref{tab:ablation}. We can observe that all the proposed components are useful to improve the privacy-preserving performance and the graph similarity measuring performance.
The variant \underline{$w/o$ Obfuscation} has poor performance. On the one hand, the obfuscated features contain fused graph-level information from both input graphs, making it difficult to infer the properties of each one. On the other hand, the obfuscation operation further helps the feature fusion between two graphs via node-graph matching, which benefits graph similarity measuring.

For the other two variants \underline{$w/o$ Context Codes} and \underline{$w/o$ N-G Matching}, we can observe that the multi-head attention techniques are vital components for developing privacy-preserved neural GSL models. Although attention layers and mean pooling are all promising graph pooling methods~\cite{xu2019gin}, the attention layers in PPGM are trained for graph similarity learning tasks, carrying less graph-level properties than the classical graph pooling method, \ie mean pooling.
The results show that the multi-head attention-based graph pooling techniques help protect user privacy when training neural graph similarity learning models.

\begin{figure}[t]
	{
		\begin{minipage}[t]{0.49\linewidth}
			\centering
			\includegraphics[width=1\textwidth]{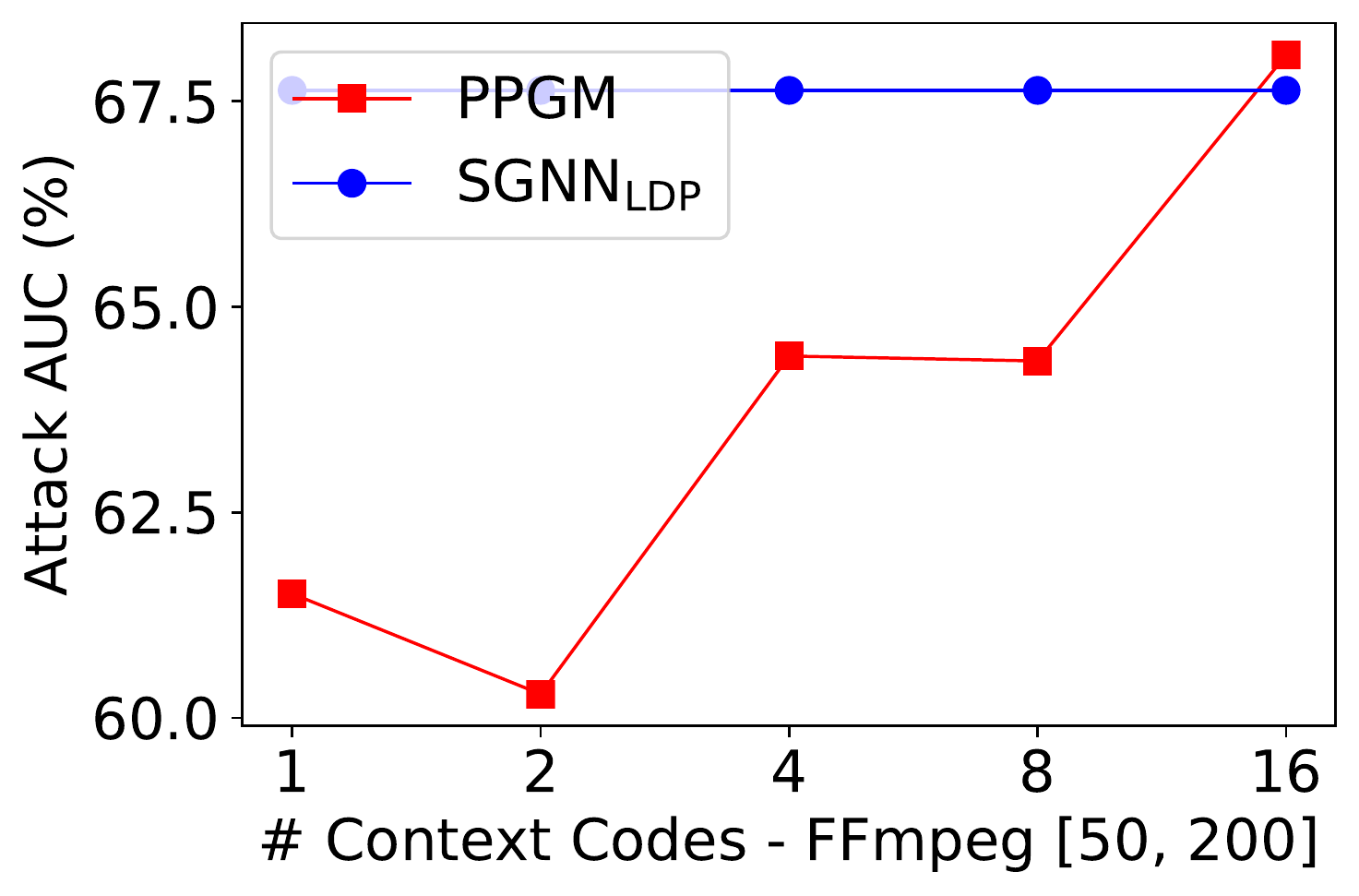}
		\end{minipage}
		\begin{minipage}[t]{0.49\linewidth}
			\centering
			\includegraphics[width=1\textwidth]{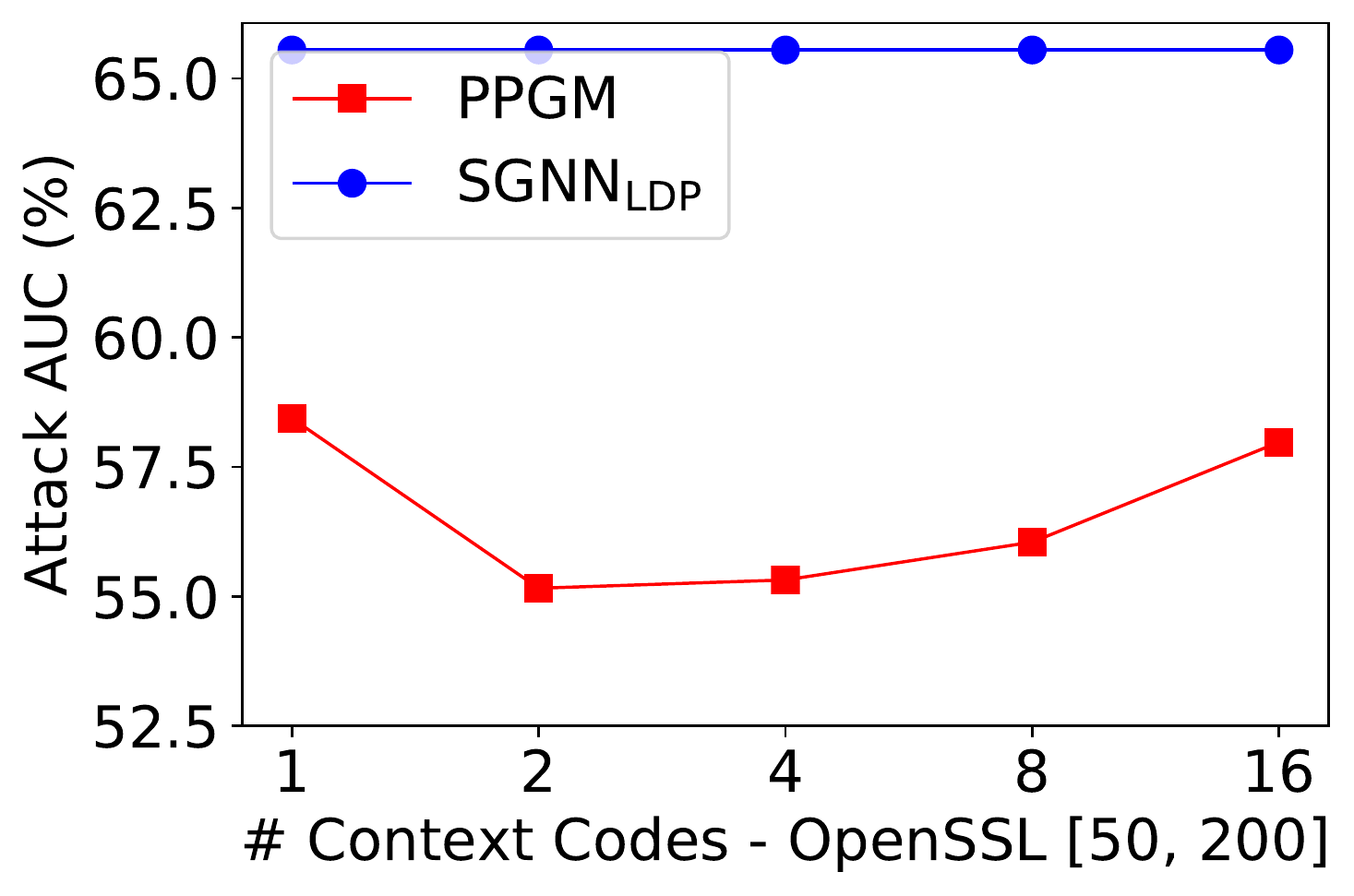}
		\end{minipage}
		\begin{minipage}[t]{0.49\linewidth}
			\centering
			\includegraphics[width=1\textwidth]{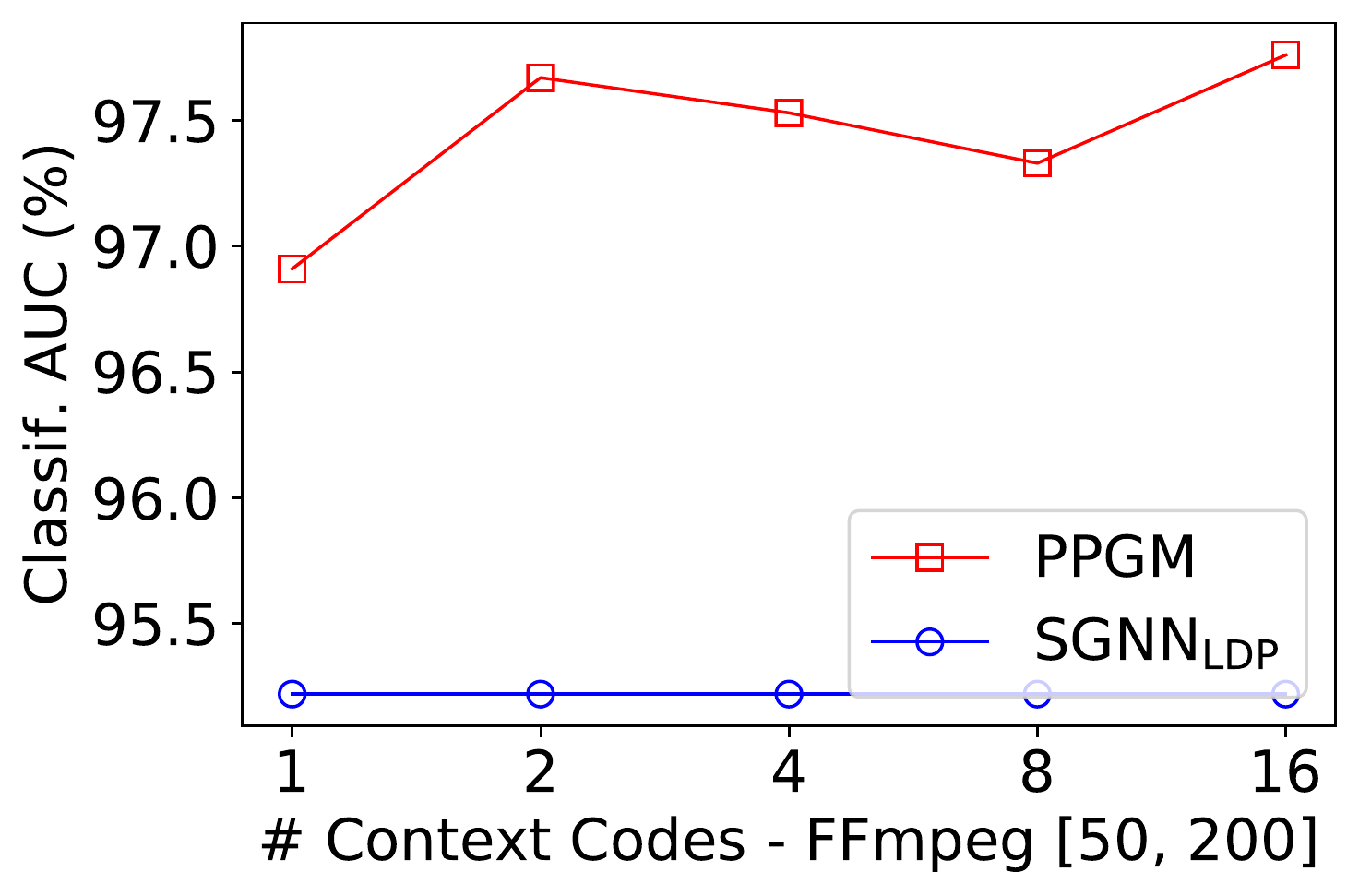}
		\end{minipage}
		\begin{minipage}[t]{0.49\linewidth}
			\centering
			\includegraphics[width=1\textwidth]{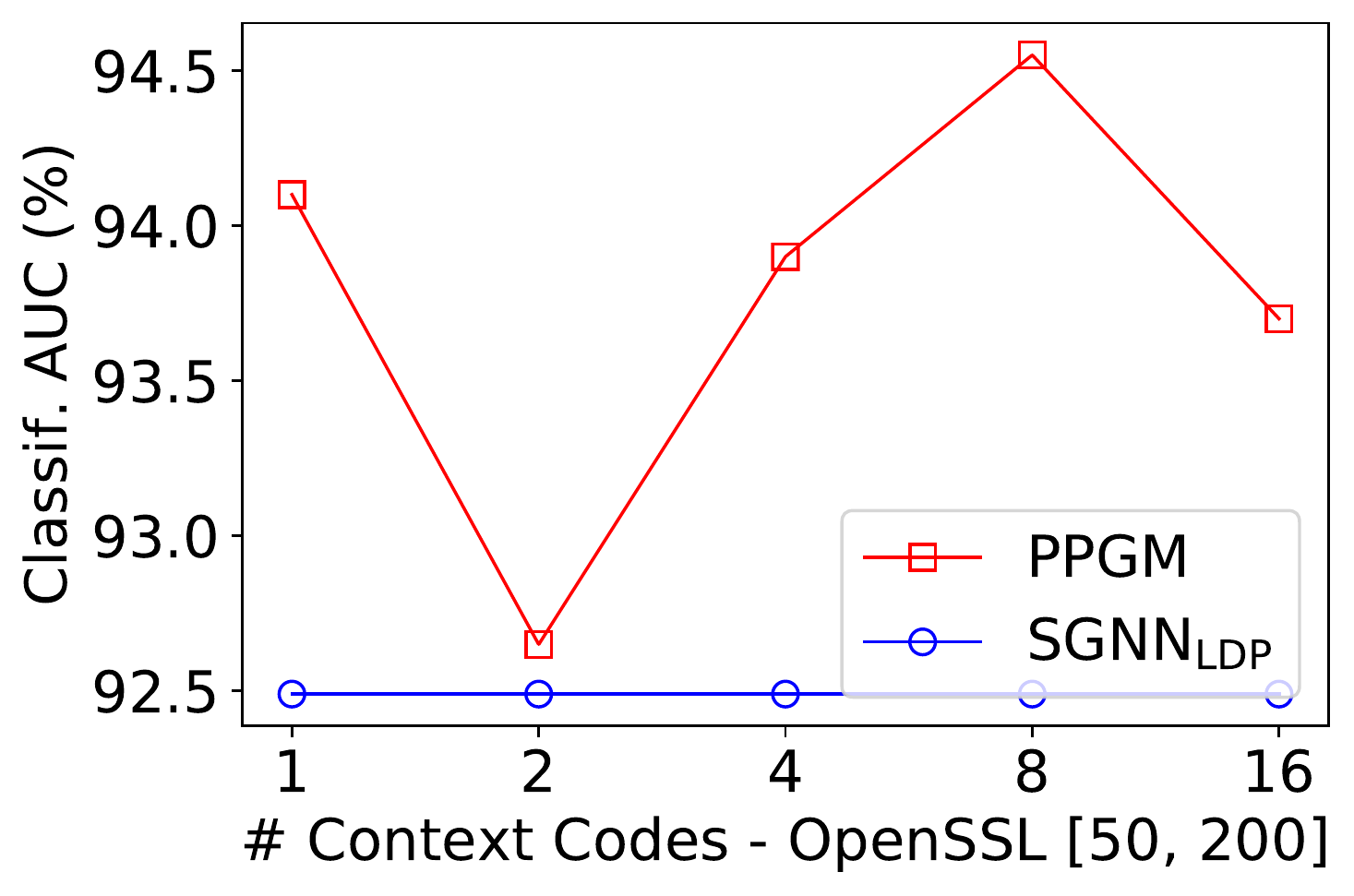}
		\end{minipage}
	}
	\caption{Performance comparison \wrt the number of context codes on the ``FFmpeg [50, 200]'' and ``OpenSSL [50, 200]'' datasets. The top shows the AUC results of the property inference attack task ($\downarrow$) and the bottom shows the AUC results of the graph similarity learning task ($\uparrow$).}
	\label{fig:tuning:n_queries}
\end{figure}

\subsection{Performance Comparison \wrt Number of Context Codes}

In the context-attentive message extraction layer defined in Eqn.~\eqref{eq:g}, the coefficient $m$ indicates the number of learnable context code vectors $\bm{c}_i$ in PPGM. The choice of $m$ indicates the perspectives of messages and correspondingly the number of obfuscated features. To analyze the influence of $m$, we vary $m\in\{1, 2, 4, 8, 16\}$ and report the results in Figure~\ref{fig:tuning:n_queries}. It shows that an appropriate choice of $m$ can improve the performance of PPGM. Specifically, when the number of context codes is too large (\ie $m\ge 16$), the  privacy-preserving performance may drop. It is probably because more context codes lead to more comprehensive  messages, which contain more user privacy properties. Overall, with different choices of $m$, PPGM outperforms the baseline method SGNN$_{\text{LDP}}$ on both the property inference attack task and graph similarity learning task. The results indicate that we need to tune this hyperparameter to find a suitable value for each dataset. Generally, we would suggest a $4 \le m \le 8$ for better privacy-preserving ability and graph similarity measuring of PPGM.

\begin{figure}[t]
	{
		\begin{minipage}[t]{0.49\linewidth}
			\centering
			\includegraphics[width=1\textwidth]{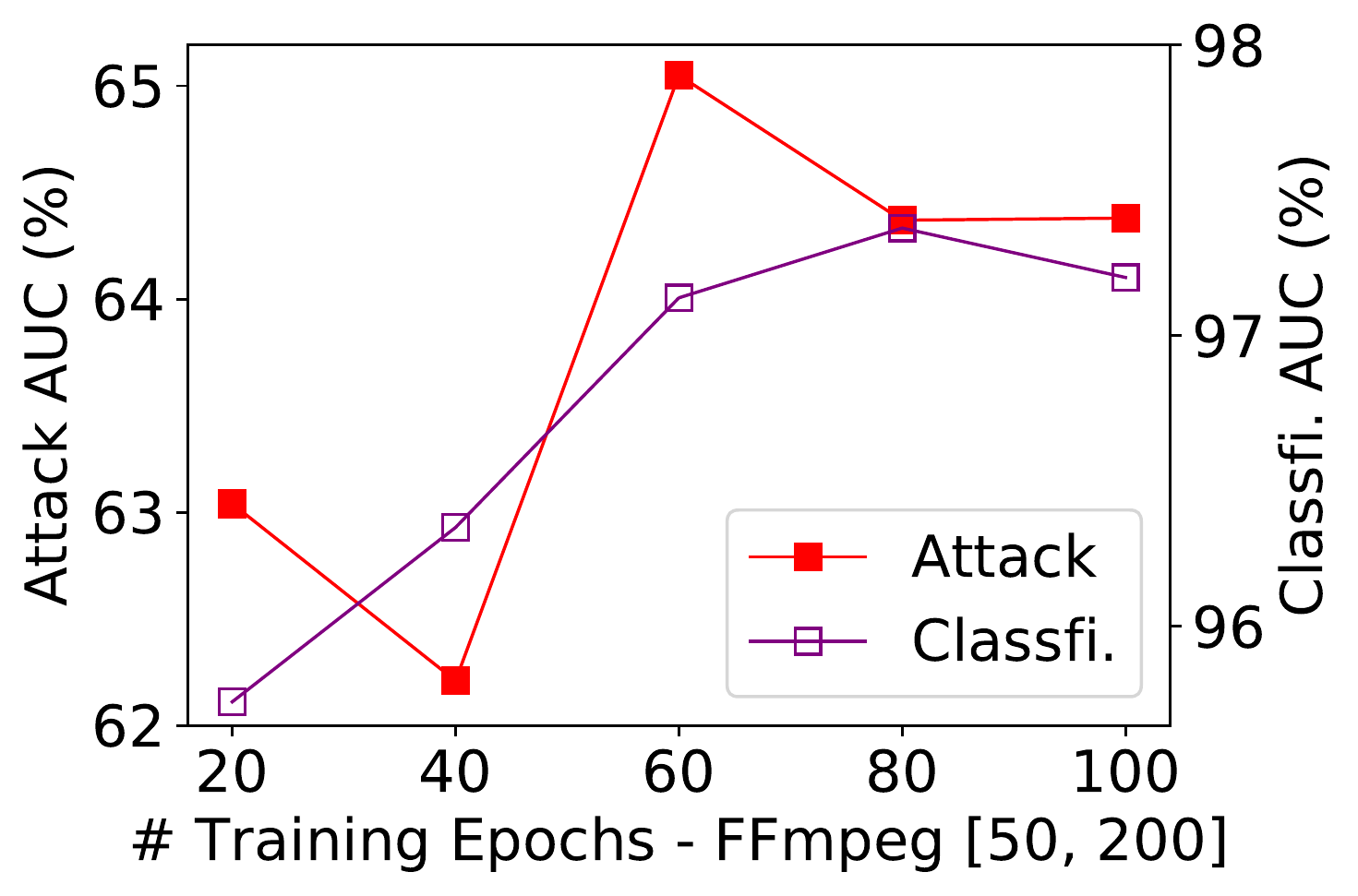}
		\end{minipage}
		\begin{minipage}[t]{0.49\linewidth}
			\centering
			\includegraphics[width=1\textwidth]{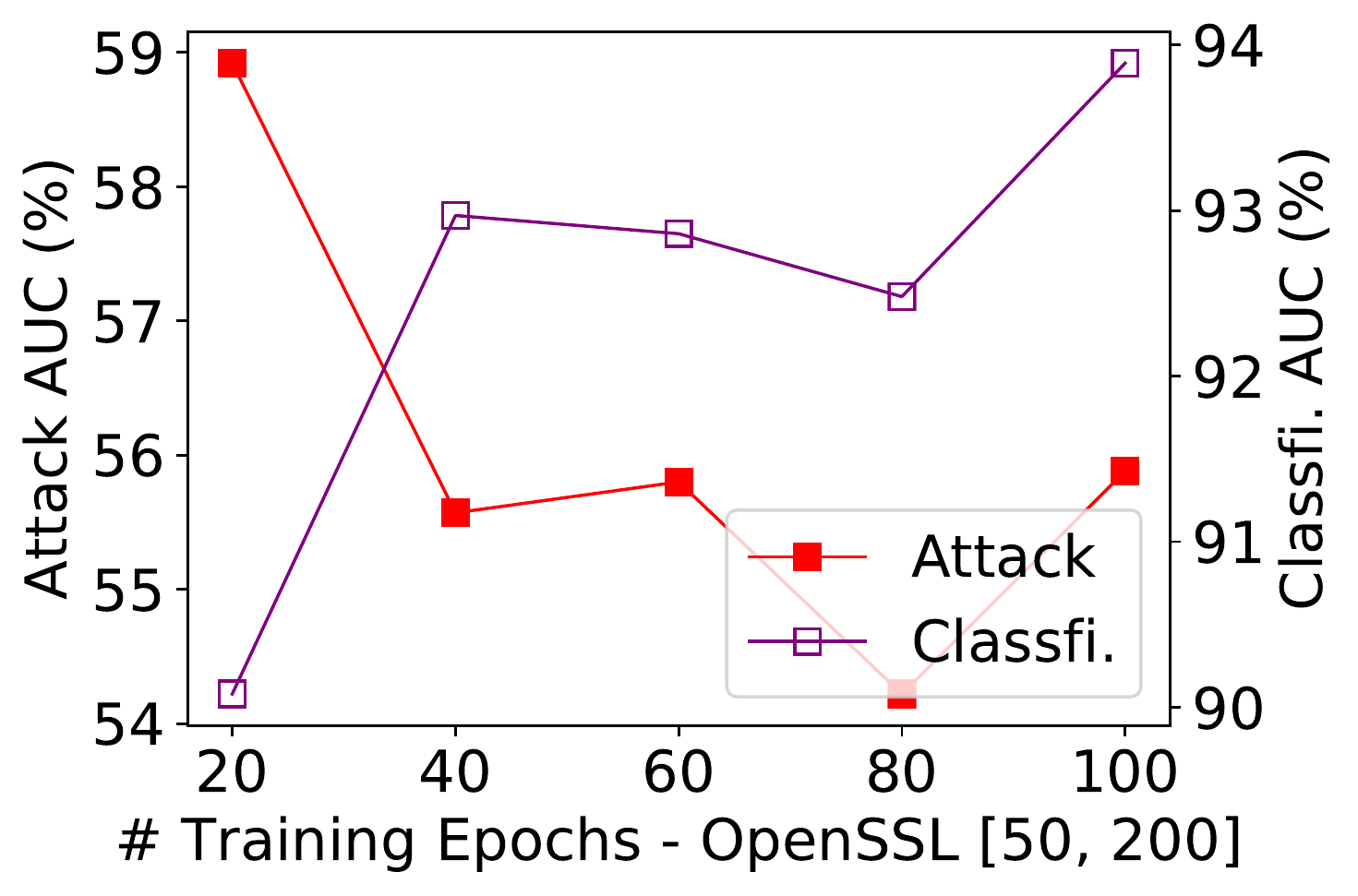}
		\end{minipage}
	}
	\caption{Performance comparison \wrt the number of training epochs on the ``FFmpeg [50, 200]'' and ``OpenSSL [50, 200]'' datasets.
The left axis shows the AUC results of the property inference attack task ($\downarrow$) and the right axis shows the AUC results of the graph similarity learning task ($\uparrow$).
	}
	\label{fig:tuning:n_epochs}
\end{figure}

\subsection{Performance Comparison \wrt Training Epochs}

For a neural graph similarity learning model that has just been initialized, the property inference attack AUC is around 0.5 (means nearly randomly classification). So it motivates us to study whether the privacy-preserving ability of PPGM may be gradually weaker with the training epochs increasing. To analyze the impact of training epochs, we vary it in the range of $\{20, 40, 60, 80, 100\}$ and show the results in Figure~\ref{fig:tuning:n_epochs}. As metrics on graph similarity measuring tasks gradually converge, the metrics on property inference attack tasks become stable. 
We do not observe significant performance drop for both the graph-graph classification task as well as the property inference attack task for even larger training epochs, \eg more than $100$ epochs.
The results show that the proposed PPGM can be trained thoroughly while preserving strong privacy protection ability.

\section{Related Work}

\subsection{Graph Similarity Learning}

Graph Similarity Learning (GSL) is a fundamental task for learning a function to quantify the similarity of two graphs~\cite{ma2021gsl_survey}.
The GSL task is widely studied in various scenarios like binary function similarity search~\cite{li2019gmn}, molecular matching~\cite{niko2017matching}, knowledge graph alignment~\cite{xu2019cross} and recommender systems~\cite{su2021gmcf}.
Conventional algorithms for graph similarity learning include Graph Edit Distance (GED)~\cite{bunke1997ged} or Maximum Common Subgraph (MCS)~\cite{bunke1998mcs}. These approaches, however, typically require exponential time complexity and cannot be applied to large-scale graphs in the real world.
Graph kernels~\cite{horv2004cyclic,yanardag2015deep_graph_kernels,niko2017matching} are also effective approaches for assessing graph similarity, but they are still memory- and time-intensive.
With the rapid growth of Graph Neural Networks (GNNs)~\cite{kipf2017gcn,velickovic2018gat,zhou2020gnn_survey}, a series of data-driven neural graph similarity learning models have been developed for improving accuracy as well as efficiency~\cite{li2019gmn,bai2019simgnn,bai2020graphsim,qin2021egsc}.
The main idea is to learn graph representations for candidate graphs with carefully designed graph matching networks~\cite{li2019gmn}. 
The key step is to introduce node-node matching scores while predicting the similarity scores.
The node-node matching scores are usually introduced via extracted histogram features~\cite{bai2019simgnn}, Convolutional Neural Networks (CNNs)~\cite{bai2020graphsim}, and inter-graph message passing techniques~\cite{li2019gmn,zhang2021h2mn,ling2021multilevel,hou2022gmpt}, serving as vital inductive biases for accurate graph similarity measuring.
Although effective, the calculation of node-node matching scores requires massive and frequent comparisons between node representations. Once deployed in privacy-sensitive scenarios, the node representations are under threat of privacy attacks, increasing the risk of sensitive data leakage and limiting the real-world application of these neural graph similarity learning models.

\subsection{Privacy of Graph Neural Networks}

Similar to images and texts, deep learning techniques for modeling graphs, \eg GNNs, rely on large-scale graph data to train effective models. Among these, there exist graph data that are collected from users that are sensitive and should be carefully leveraged, such as personal user portraits~\cite{su2021gmcf}, healthcare data~\cite{li2020diagnosis}, and bioinfomatics~\cite{niko2017matching,li2021braingnn}. 
Various recent studies show that vanilla graph neural networks may suffer from privacy attacks, including membership inference~\cite{duddu2020quantifying,olatunji2021membership,he2021node,wu2021adapting}, property inference~\cite{zhang2022inference}, and reconstruction attack~\cite{he2021stealing,zhang2021graphmi}.
Facing the threats, effects have been made for developing trustworthy GNNs~\cite{dai2022comprehensive}.
A series of techniques are proposed, such as differential privacy~\cite{olatunji2021releasing,xu2018dpne,zhang2021graph,sajadmanesh2021locally}, federated learning~\cite{wu2021fedgnn,wang2020graphfl,pei2021decentralized,xie2021federated,zheng2021asfgnn}, and adversarial learning~\cite{li2021adversarial,liao2021information,wang2021privacy}.
However, existing work about GNN privacy-preserving consider tasks like node classification, link prediction, and graph classification, which take only one graph as input. The emerged privacy issues of calculating graph similarity have not been properly investigated. Especially, neural GSL models are frequently deployed in privacy-sensitive scenarios like binary function similarity search~\cite{li2019gmn}, bioinfomatics~\cite{niko2017matching}, and recommender systems~\cite{su2021gmcf}.

\section{Conclusion}
In this paper, we study the privacy protection ability of neural graph similarity learning models.
Based on the proposed conception of \emph{attackable representations}, we systematically summarize the privacy attacks a neural GSL model may suffer from.
We then correspondingly propose an effective and privacy-preserved model, named \emph{PPGM}.
To prevent reconstruction attacks, we do not send node representations off the user device. To alleviate the attacks to graph properties, obfuscated features that contain information on both graphs are communicated between devices. Effective node-graph matching techniques and multi-perspective graph representations help improve the similarity measuring performances. We further propose to quantitatively evaluate the privacy-preserving ability of neural graph similarity learning models via training black-box attack models.
For future work, we will study how different choices of neural network architectures influence the privacy-preserving ability. We will also study the design space of neural GSL models under consideration of privacy-preserving~\cite{you2020design_space,wang2022space_gnn_cf}. Besides, we will explore how to generalize the proposed model to more complicated graph structures, \eg knowledge graphs and heterogeneous graphs.

\section{Acknowledgements}
This work was partially supported by National Natural Science Foundation of China under Grant No. 61872369,
Beijing Natural Science Foundation under Grant No. 4222027,  and Beijing Outstanding Young Scientist Program under Grant No. BJJWZYJH012019100020098.
This work was partially supported by Beijing Academy of Artificial Intelligence~(BAAI).
Yupeng Hou was also partially supported by Alibaba Group through Alibaba Research Intern Program.
Xin Zhao is the corresponding author.
\balance

\clearpage
\bibliographystyle{IEEEtran}
\balance
\bibliography{main}

\end{document}